\documentclass{article}

\usepackage[preprint]{neurips_2026}

\usepackage[utf8]{inputenc} 
\usepackage[T1]{fontenc}    
\usepackage{hyperref}       
\usepackage{url}            
\usepackage{booktabs}       
\usepackage{amsfonts}       
\usepackage{nicefrac}       
\usepackage{microtype}      
\usepackage{xcolor}         
\usepackage{graphicx}
\usepackage{placeins}
\usepackage{amsmath}
\usepackage{algorithm}
\usepackage{algpseudocode}
\usepackage{wrapfig}
\usepackage{xspace}

\title{Zero-shot Imitation Learning by \\ Latent Topology Mapping}

\newcommand{\zalt}{ZALT\xspace}
\newcommand{\zalts}{ZALT's\xspace}

\author{%
  Maxwell J. Jacobson,   Yexiang Xue \\
  Purdue University, Department of Computer Science\\
  West Lafayette, IN\\
  \texttt{\{jacobs57, yexiang\}@purdue.edu} \\
}

\begin{document}

\maketitle

\begin{abstract}
Imitation learning is effective for training agents when expert demonstrations are available, but collecting demonstrations for every complex task in an environment is costly. We study the long-horizon, goal-conditioned setting where a fixed demonstration dataset contains useful behavior, but not complete examples for every task the agent must solve. Existing imitation learning methods can learn strong policies from demonstrations, but when solving long-horizon tasks, small errors accumulate over long primitive-action trajectories and make zero-shot adaptation to new tasks unreliable. We introduce Zero-shot Agents from Latent Topologies (ZALT), an imitation-learning method that solves unseen start--goal tasks beyond those demonstrated during training. \zalt identifies latent hub states where trajectories converge or diverge, learns policies and a dynamics model over hub-to-hub transitions, and plans over the hub topology to complete new tasks. This topology makes demonstrated behaviors explicitly composable while compressing long tasks into shorter sequences of abstract transitions -- combined, these enable \zalt to perform zero-shot adaptation. In a complex 3D maze environment, \zalt achieves 55\% zero-shot success on unseen tasks, compared to 6\% for the strongest baseline.
\end{abstract}

\section{Introduction}

\begin{figure}[thb]
    \centering
    \includegraphics[width=1\linewidth]{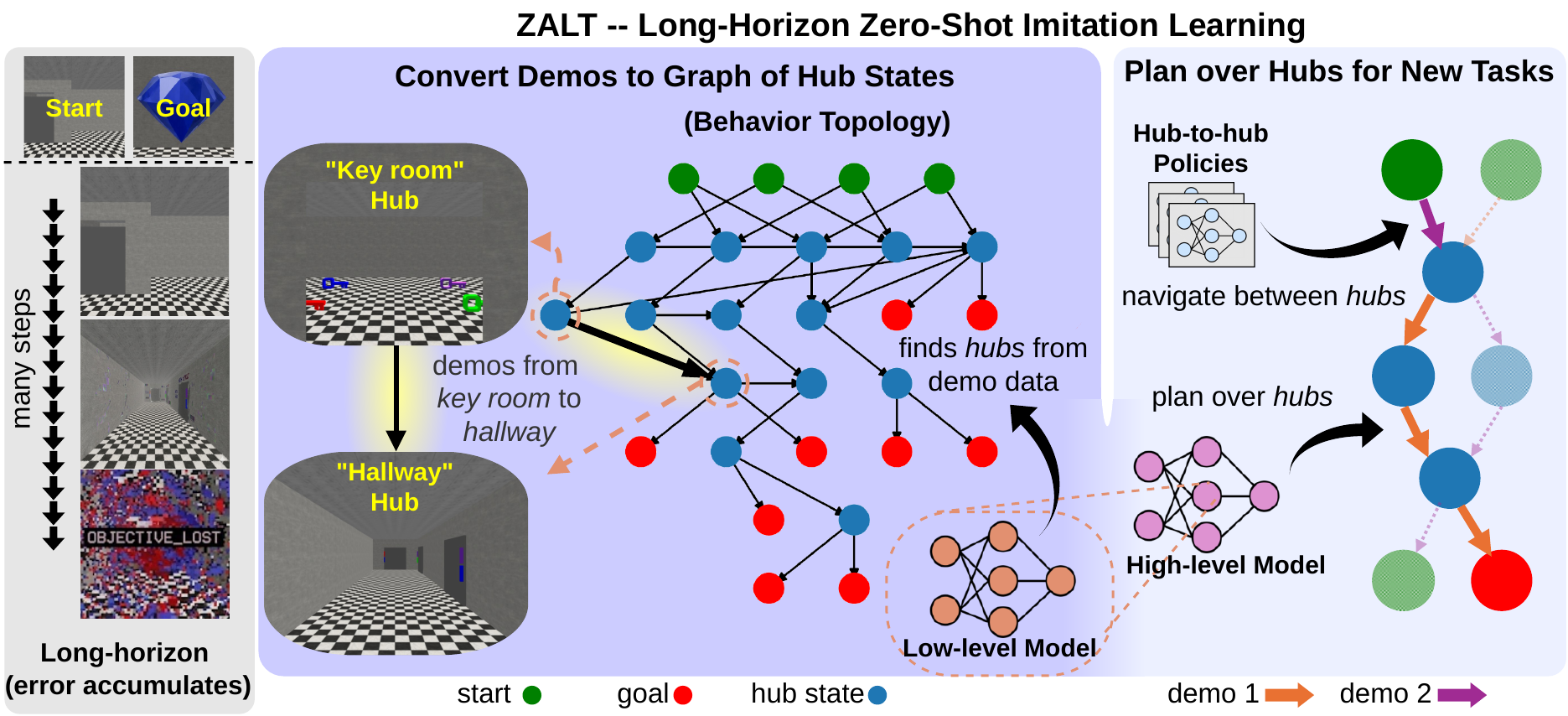}
    \vspace{-5pt}
    \caption{Directly composing behavior at the primitive-action level requires many sequential decisions, allowing small policy, model, or distribution-shift errors to compound until the objective is lost (\textbf{left}). ZALT instead learns a behavior topology from demonstrations, using a low-level dynamics model to identify a few reusable hub states (\textbf{middle}). Then, a high-level dynamics model plans over those hubs, and demonstration-trained policies execute the selected hub-to-hub segments (\textbf{right}). This compresses a long primitive horizon into a shorter sequence of abstract transitions, enabling zero-shot composition of new start–goal behaviors with fewer opportunities for error accumulation. Demonstration video: \url{https://arcosin.github.io/Zalt/}}
    \label{fig:hymap_main}
    \vspace{-13pt}
\end{figure}

Imitation learning is a powerful technique for training agents from expert behavior, especially when direct reward design or online exploration is difficult. We study imitation learning in long-horizon, goal-conditioned environments where a fixed demonstration dataset covers some start--goal tasks, but the agent must solve new tasks \textit{zero-shot}, without additional interaction or training. This setting captures a central promise of imitation learning: demonstrations should teach more than isolated task instances, since collecting expert trajectories for every possible long-horizon task combination is expensive or impossible in interactive agents, game environments, and robotics.

Prior imitation-learning methods provide strong ways to learn from expert behavior. Behavioral cloning and goal-conditioned imitation can learn effective policies directly from demonstrations \cite{hussein2017imitation,osa2018algorithmic}, while sequence models \cite{chen2021decision} and model-based imitation methods \cite{hu2022model,bronstein2022hierarchical,kidambi2021mobile} can use demonstration data to predict or compose behavior beyond a single observed trajectory. However, zero-shot generalization becomes much harder in long-horizon tasks: the agent must compose behavior outside the demonstrated start--goal distribution, and policy, model, or distribution-shift errors have hundreds of steps over which to accumulate \cite{ross2011dagger}. As a result, a fixed demonstration dataset can contain the local behavior needed for a new task while still failing to support reliable full-task execution.

We introduce Zero-shot Agents from Latent Topologies (ZALT), an imitation-learning method for zero-shot long-horizon behavior composition. \zalt processes demonstrations into a latent \textit{behavior topology} by identifying \textit{hub states} where trajectories converge or diverge. These hubs act as reusable junctions in the demonstration data. Edges in the topology correspond to demonstrated multi-step behavioral segments between hubs, allowing long tasks to be represented as shorter sequences of abstract transitions (see Figure~\ref{fig:hymap_main}, middle panel). \zalt then learns how to use the behavior topology. A high-level dynamics model predicts likely next hubs from the sequence of hubs visited so far, capturing path-dependent information that may not be reliable in a single low-level latent state over an entire long-horizon task. Separately, demonstration-trained policies learn to execute hub-to-hub transitions using the trajectory segments associated with each topology edge. At test time, \zalt matches the start and goal to hubs, searches for a most plausible route through the topology, and executes that route by chaining the learned hub policies (see Figure~\ref{fig:hymap_main}, right panel). 

\zalt addresses long-horizon composition by performing composition at the hub level rather than the primitive-action level. This compresses a long task into a shorter sequence of abstract transitions, reducing opportunities for error accumulation while still grounding execution in policies learned from expert demonstrations. As a result, \zalt can solve new start--goal tasks zero-shot after processing a fixed demonstration dataset.

We evaluate \zalt in a complex 3D maze environment requiring long-horizon ordered object manipulation with keys, locked doors, diamonds, and a barrel. The key challenge is zero-shot composition: the agent must solve held-out start--goal tasks whose complete successful trajectories never appear in the demonstration data. \zalt achieves 55\% zero-shot success on unseen tasks, compared to 6\% for the strongest baseline and 0\% for the remaining baselines. It also achieves 72.2\% success on seen tasks, while the strongest baseline reaches only 11\%, showing that the task is difficult even before held-out generalization. Successful \zalt plans compress roughly 244--281 primitive timesteps into about 33 hub-level transitions, with each abstract edge encoding about 7.9 primitive actions on average. We also collect a video of a successful zero-shot \zalt execution, shown in Figure~\ref{fig:hymap_live}, illustrating how the learned topology is used during live inference.


\section{Problem Definition}

\zalt is designed to solve long-horizon tasks by composing behaviors learned from an incomplete set of expert demonstrations, enabling new tasks to be solved zero-shot. We formalize this setting as a fixed environment with many possible start--goal tasks. Each task is specified by an initial state ($s_0$) and a goal ($g$). The agent has access to an offline demonstration dataset collected from a subset of possible start--goal combinations, and must solve new start--goal combinations without additional environment interaction or policy optimization. Let $\mathcal{S}$ be the state space, $\mathcal{A}$ be the action space, and $\mathcal{G}$ be the goal space. A task is a pair $(s_0,g) \in \mathcal{S}_0 \times \mathcal{G}$, where $\mathcal{S}_0 \subseteq \mathcal{S}$ is the set of possible initial states. Some imitation learning formulations define the problem over a Markov Decision Process (MDP), and therefore also define a reward function $r(s, a, g)$.

The agent receives a fixed dataset of expert demonstrations $\mathcal{D} = \{\tau_i\}_{i=1}^{N}$. Each trajectory is written as $\tau_i = (s^i_0, a^i_0, s^i_1, a^i_1, \ldots, s^i_{T_i}, g_i, y_i)$, where $g_i$ is the intended goal and $y_i \in \{0,1\}$ indicates whether the trajectory successfully achieved that goal. We allow $y_i=0$ because expert-generated demonstrations may still fail in long-horizon environments, and failed trajectories can provide useful information about behavioral connectivity and unsupported routes. 

The objective is zero-shot behavior composition: use the structure present in $\mathcal{D}$ to solve unseen start--goal combinations by recombining demonstrated behavioral segments. $\mathcal{D}$ contains only some possible start--goal pairs ($s_0, g$). At test time, the agent is evaluated on held-out start--goal pairs that do not appear in $\mathcal{D}$. A held-out pair may still involve familiar starts, goals, objects, rooms, or local behaviors, but the full task from that specific start to that specific goal was never demonstrated during training. 

A method succeeds on a test task if the executed trajectory reaches the specified goal within the environment horizon.  This setting assumes that useful behavioral connectivity exists in the demonstration data. A complete test trajectory may not be present, but portions of the required behavior may appear across different demonstrations. The central problem is to learn this structure and compose it into a valid policy for a new task.

\section{Zero-shot Agents from Latent Topologies (ZALT)}

\zalt delivers zero-shot behavior in the long-horizon imitation-learning setting by converting a fixed set of expert demonstrations into a composable, high-level structure for planning. We define a \textit{behavior topology} as a graph-like abstraction of the demonstration data, where nodes are latent \textit{hub states} and edges represent demonstrated multi-step behavior segments between hubs. \zalt constructs this topology by identifying hubs in the latent trajectories: regions where demonstrations converge into the same latent area or diverge toward different future behaviors. Because observations are high-dimensional, \zalt detects these hubs in a learned latent space obtained from a low-level encoder--decoder latent dynamics model whose prediction objective encourages behaviorally similar states to have nearby representations. After constructing the topology, \zalt learns how to navigate it. A high-level dynamics model predicts movement over hubs rather than primitive states, while demonstration-trained policies learn to execute hub-to-hub transitions using the corresponding trajectory segments. At inference time, given an unseen start--goal pair, \zalt plans over the behavior topology using the high-level model to guide search toward likely successful hub sequences. The resulting abstract plan is then converted into low-level control by executing the learned hub-to-hub policies. In this way, \zalt learns explicitly composable behaviors from demonstrations while compressing long primitive-action horizons into shorter sequences of symbolic hub-to-hub transitions.

\subsection{Finding Hubs with Convergences and Divergences}

The first step in \zalt is to build the behavior topology by finding hub states in the demonstration data. Hubs are latent states that can connect many possible start--goal tasks. 
Without hubs, a long-horizon task must be solved as a long sequence of primitive actions, which gives many opportunities for small errors to accumulate. 
With hubs, the same task can be treated as a shorter sequence of learned behaviors. In this sense, hubs act as junctions: they are states where different demonstrations meet, separate, begin, or end, and therefore provide useful places to compose behavior, while isolating errors to the links between them.

We identify two main kinds of hub states: convergences and divergences. A \textit{convergence} is a state that can be reached from multiple distinct prior states or trajectories. For example, if several demonstrations enter the same hallway intersection from different rooms, that intersection is a natural convergence point. A \textit{divergence} is a state from which multiple distinct future behaviors are possible. For example, if the agent reaches the same hallway intersection and then different demonstrations continue toward different doors, that intersection is a natural divergence point. Convergences identify broadly reachable states, while divergences identify meaningful decision points. Together, they capture the reusable structure needed for long-horizon behavior composition.

\begin{wrapfigure}{r}{0.4\textwidth}
    \centering
    \includegraphics[width=0.35\textwidth]{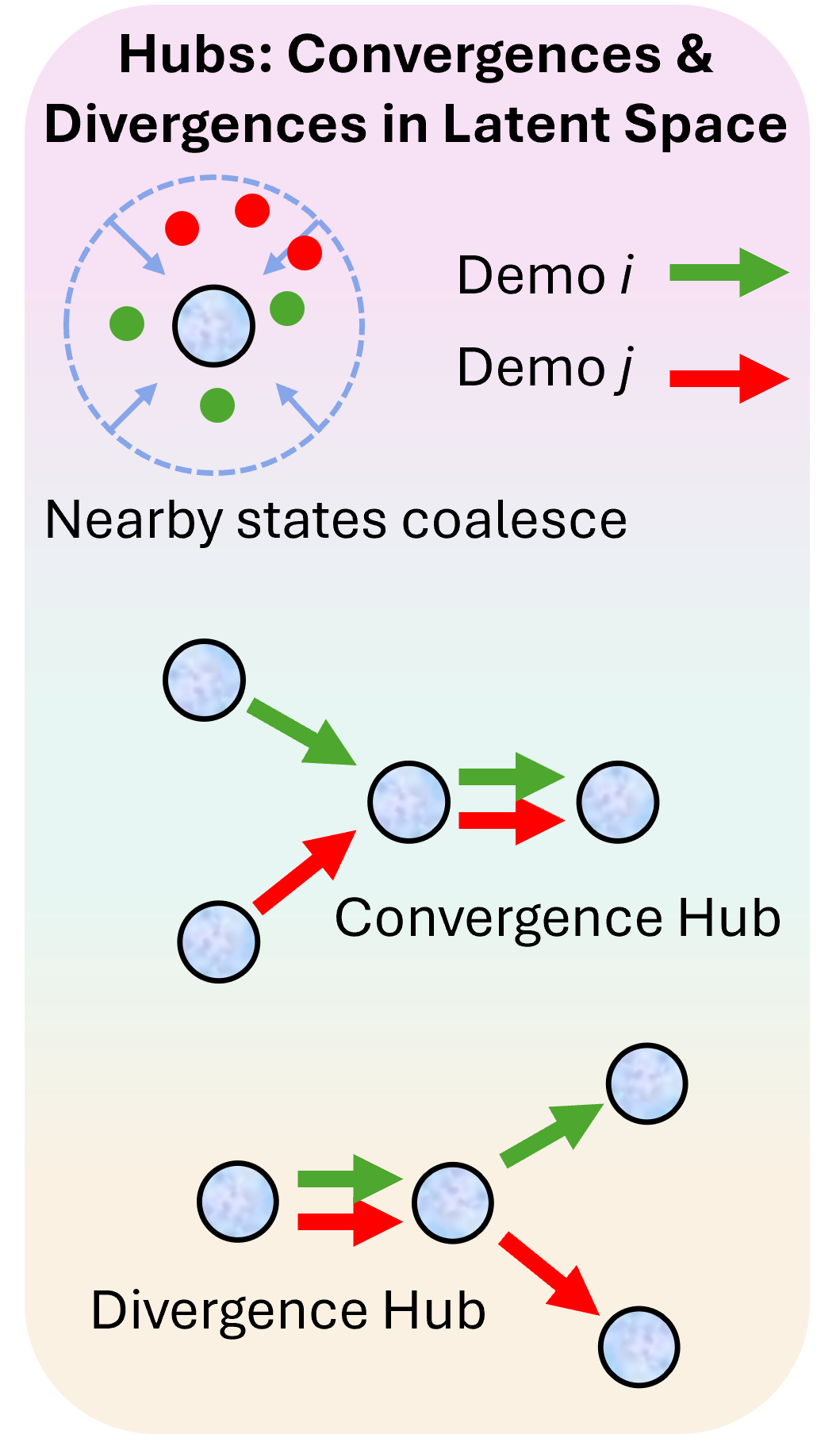}
    \caption{\textbf{Hub formation.} \zalt coalesces demonstration latent vectors into clusters. Hubs are clusters where demos converge from multiple previous clusters or diverge into multiple next clusters.}
    \label{fig:converge_diverge}
    \vspace{-4pt}
\end{wrapfigure}

In practice, exact state equality is too strict, especially with high-dimensional observations. \zalt therefore detects convergences and divergences in a learned latent space (see Figure~\ref{fig:converge_diverge}). We assume each observed state $s_t$ has been encoded into a dynamics-aligned latent state $z_t$, so each demonstration becomes a latent trajectory. We write the encoded demonstrations as $\mathcal{T}_z = \{\zeta^1,\ldots,\zeta^N\}$, where $\zeta^i = (z^i_0, a^i_0, z^i_1, a^i_1, \ldots, z^i_{T_i})$. Since latent states are continuous, we group nearby latent states using a tolerance $\epsilon$. Let $K_\epsilon(z)$ be the tolerance cluster containing latent state $z$. A cluster $K$ is a convergence candidate if it is reached from multiple distinct previous clusters: $|\{K_\epsilon(z^i_{t-1}) : K_\epsilon(z^i_t)=K\}| \geq 2$. A cluster $K$ is a divergence candidate if it leads to multiple distinct next clusters: $|\{K_\epsilon(z^i_{t+1}) : K_\epsilon(z^i_t)=K\}| \geq 2$. We also include demonstration start states and terminal states as hubs, since starts provide entry points into the topology and terminal states represent task outcomes.


Because hub discovery depends on latent similarity, the latent representation must emphasize behaviorally meaningful structure. A representation based only on visual reconstruction may group states that look alike but lead to different outcomes, or separate states that look different but support the same next behaviors. \zalt therefore learns a dynamics-aligned latent space: states should be represented in a way that supports predicting what actions can do next and what recent behavioral context led to the current state. This makes convergence and divergence detection more likely to identify reusable behavioral junctions rather than superficial observation matches.

To learn this representation, \zalt uses a low-level encoder--dynamics--decoder model over primitive actions, following the general latent-dynamics and world-model approach used in model-based reinforcement learning \cite{ha2018world,hafner2019planet,hafner2020dreamer}. The encoder ($Enc_{\theta_e}(s_t,\text{history}_t) \rightarrow z_t$) maps the current observation and recent history into a compact latent state, the low-level dynamics model ($M^\text{low}_{\theta_m}(z_t,a_t) \rightarrow \hat{z}_{t+1}$) predicts the next latent state after a primitive action, and the decoder ($Dec_{\theta_d}(\hat{z}_{t+1}) \rightarrow \hat{s}_{t+1}$) reconstructs next-state information from the predicted latent. The purpose is not to use this model as a long-horizon planner, but to learn a dynamics-aware latent space where behaviorally similar states can merge and useful hubs can emerge. We use a GRU \cite{cho2014gru} memory module with clipped history so the latent can capture recent context without memorizing full trajectory identity. Additional implementation details are provided in Appendix~\ref{sec:appendix_low_level_dynamics}.

\subsection{Learning a Hub Dynamics Model}
\label{sec:method_high_model}

After finding hubs, \zalt learns a high-level dynamics model over the behavior topology. Although each hub latent is conditioned on the current state and recent memory, \zalt does not rely on that memory to remain reliable across an entire long-horizon task. Instead, the important behavioral history is represented explicitly as the sequence of hubs visited so far. The hub dynamics model learns how this hub history predicts the next hub. This is necessary because the current hub alone may not determine what should happen next. For example, two demonstrations may both pass through the same hallway hubs before approaching a door, but whether the door can be opened depends on the prior behavior: did the agent previously collect a compatible key, or did it collect the wrong one? A raw graph edge may say that the hallway can lead toward the door, but the hub dynamics model learns whether that transition is likely to lead to a successful opened-door hub or to a failed/unsupported outcome given the previous hub sequence.

We train this model from hub sequences extracted from demonstrations. Each demonstration is encoded into latent states, matched to hubs, and collapsed into a sequence of distinct hub visits. From a sequence such as $(h_0,h_1,h_2,h_3)$, \zalt creates supervised examples $(h_0) \mapsto h_1$, $(h_0,h_1) \mapsto h_2$, and $(h_0,h_1,h_2) \mapsto h_3$. We write the hub history up to abstract time $m$ as $h_{0:m}=(h_0,\ldots,h_m)$. The high-level dynamics model is a GRU over hub embeddings followed by a softmax classifier, $M^\text{high}_{\theta_h}(h_{0:m}) \rightarrow P(h_{m+1} \mid h_{0:m})$. It is trained with cross-entropy loss on the demonstrated next hub, equivalently minimizing $-\log P(h_{m+1} \mid h_{0:m})$ over the hub-sequence examples.

The prediction is masked by the edges of the topology. If no demonstration ever contains a segment from hub $h_i$ to hub $h_j$, then edge $(h_i,h_j)$ is absent, and the transition is assigned zero probability. This is important because \zalt has no demonstrated segment from which to train a low-level policy for executing such a transition. Thus, $M^\text{high}_{\theta_h}$ only predicts among outgoing neighbors supported by the demonstration data.


\subsection{Learning Policies Between Hubs}

After constructing the behavior topology, \zalt learns policies for executing its edges. These essentially convert a high-level subtask like ``go from hub 7 to 8'' into low-level control. Many parameterizations are possible, but in this work, we use one policy per source hub, conditioned on the target hub embedding. For edge $(h_i,h_j)$, policy $\pi_{h_i}$ predicts primitive actions as $\pi_{h_i}(a_t \mid s_t,h_j)$. Each source-hub policy has its own GRU memory, allowing it to use local execution history while traversing a hub-to-hub segment.

Training data is taken directly from demonstrated topology edges. When a demonstration visits $h_i$ and next reaches $h_j$, the intervening primitive actions become a behavior-cloning segment for edge $(h_i,h_j)$. All outgoing-edge segments from $h_i$ train $\pi_{h_i}$, with the target hub embedding specifying which successor hub to reach. Policies are trained with cross-entropy on demonstrated actions.

To make these policies less brittle at hub boundaries, we use stochastic boundary perturbation during training. With high probability, the learner receives the canonical hub-to-hub segment; otherwise, it receives either a truncated segment, which removes early context, or an extended segment, which adds a short approach trajectory, encouraging the policy memory to act correctly under perturbed boundary conditions. In our implementation, this is combined with small observation noise and action-label smoothing \cite{szegedy2016rethinking,muller2019when}. These regularizers make the policies more robust to slight timing errors in hub detection and execution.

\subsection{Zero-shot Inference}

With the behavior topology, high-level hub dynamics model, and hub-to-hub policies learned, \zalt can solve unseen start--goal pairs without additional training. Essentially, \zalt uses the high-level model to search for a plausible path through the topology, then executes the resulting plan using the learned hub-to-hub policies. Given a start observation $s_0$, \zalt encodes it into a latent state and matches it to a start hub $h_s$. Given a goal $g$, \zalt identifies the set of terminal success hubs $\mathcal{H}_g$ whose stored goal labels match $g$. Inference then consists of two steps: first search for the most plausible hub route from $h_s$ to some $h_g \in \mathcal{H}_g$, then execute that route by chaining the corresponding hub policies.

The search is performed over hub histories rather than individual hubs. A partial route is written as $H_m=(h_0,\ldots,h_m)$, with $h_0=h_s$. For each partial history, the high-level dynamics model estimates $M^\text{high}_{\theta_h}(H_m) = P(h_{m+1}\mid H_m)$. This lets the search use path-dependent information: the plausibility of moving from the current hub to a next hub can depend on the previous hubs, not only on the current hub. Transitions absent from the behavior topology are masked out (see Section \ref{sec:method_high_model}), so $P(h'\mid H_m)=0$ whenever $(h_m,h')$ is not an edge. Low-probability transitions are also rejected using a probability floor $p_{\min}$.

Among valid hub histories, \zalt selects a goal-reaching route using a \textit{bottleneck transition cost}. Extending $H_m$ with a candidate next hub $h'$ gives $H_{m+1}=(h_0,\ldots,h_m,h')$ and cost $C(H_{m+1})=\max(C(H_m),-\log P(h'\mid H_m))+\eta$, where $\eta$ is a small hop penalty. This objective prefers routes whose weakest hub transition is still likely, while mildly favoring shorter routes when bottleneck quality is similar. The selected history induces a hub-edge plan $((h_0,h_1),(h_1,h_2),\ldots)$.

Execution chains the learned hub policies. For each planned edge $(h_i,h_j)$, \zalt selects the source-hub policy $\pi_{h_i}$ and conditions it on the target hub embedding $h_j$. The policy emits primitive actions until the agent reaches the expected next hub $h_j$, at which point \zalt advances to the next edge. 

For example, suppose one demonstration contains the hub sequence $(h_A,h_E,h_F,h_G,h_H)$ and another contains $(h_B,h_E,h_D,h_K,h_P)$. A new task may require starting near $h_B$ and reaching a goal associated with $h_H$, even though no demonstration shows the full route $(h_B,\ldots,h_H)$. If the topology contains the shared hub $h_E$ and the high-level model assigns sufficient probability to the history $(h_B,h_E,h_F,h_G,h_H)$, \zalt can compose the prefix from the second demonstration with the suffix from the first. The resulting primitive behavior is produced by executing the corresponding policies for $(h_B,h_E)$, $(h_E,h_F)$, $(h_F,h_G)$, and $(h_G,h_H)$.




\section{Related Work}

\zalt addresses long-horizon, zero-shot behavior in goal-conditioned imitation learning from incomplete demonstrations. It is most closely related to imitation learning and offline RL, which learn behavior from fixed data. \zalt also connects to hierarchical reinforcement learning and other methods that learn reusable skills, since it decomposes long tasks into shorter behavioral segments. These areas address pieces of the problem, but \zalt targets the specific challenge of composing new long-horizon task solutions zero-shot from demonstrated behavior.

Imitation learning and offline reinforcement learning are central to \zalt because both learn from fixed datasets, but they use those datasets differently. Imitation learning treats demonstrations as behavior to reproduce, using methods that clone expert actions, correct learner-induced distribution shift, infer rewards, or match expert behavior distributions \cite{hussein2017imitation,osa2018algorithmic,ng2000algorithms,ho2016generative,fu2018learning}. Some methods, including \zalt, make use of learned dynamics models in this setting \cite{hu2022model,bronstein2022hierarchical,kidambi2021mobile}, using expert behavior to extrapolate how the environment itself operates. Goal- and outcome-conditioned imitation extend this idea by conditioning policies on desired goals, returns, or outcomes, allowing a single policy to imitate many demonstrated behaviors \cite{liu2022goal,emmons2022rvs,chen2020bail,wang2020critic}. Offline RL instead treats the fixed dataset as experience for estimating which actions maximize reward, often using conservative or policy-constrained objectives to avoid unsupported actions, or learned models and sequence representations to plan from data \cite{levine2020offline,prudencio2023survey,fujimoto2019off,kumar2019stabilizing,wu2019behavior,kumar2020conservative,kostrikov2022offline,kidambi2020morel,yu2020mopo,yu2021combo,chen2021decision,janner2021offline}. \zalt builds on these ideas by targeting long-horizon composition, where even methods that can generalize demonstrated behavior may fail because small policy, value, or model errors compound across many primitive decisions. \zalt still learns from demonstration data, but reduces this failure mode by composing behavior over shorter hub-to-hub transitions.

\zalt also connects to hierarchical reinforcement learning (HRL), since it decomposes long-horizon behavior into high-level decisions over reusable segments and low-level policies that execute those segments \cite{pateria2021hierarchical}. Subgoal and feudal HRL methods learn high-level policies that choose intermediate objectives and low-level policies that reach them, shortening the effective planning horizon and often giving agents denser intermediate structure \cite{dayan1993feudal,kulkarni2016hierarchical,nachum2018data,levy2019learning,vezhnevets2017feudal}. Options and policy-tree methods similarly use temporally extended actions, allowing a high-level controller to select skills that compress many primitive actions into one decision \cite{sutton1999between,dietterich2000hierarchical,bacon2017option,zhang2019dac}. Skill and subtask discovery methods learn reusable behaviors, bottlenecks, or graph-like abstractions that can support exploration, transfer, or high-level planning \cite{konidaris2009skill,mcgovern2001automatic,simsek2008skill,menache2002q,achiam2018variational,co-reyes2018self,eysenbach2019search,sohn2018hierarchical, shi2022skimo}. These methods provide the key insight that long-horizon behavior becomes more tractable when it is decomposed into reusable pieces, but many rely on task-specific interaction, rewards, or adaptation to learn which high-level sequence solves a new task. Meta-RL is also related because it studies reusable knowledge across tasks, but it usually emphasizes fast adaptation through learned policies, initializations, or latent contexts \cite{beck2023survey,duan2016rl2,wang2016learning,finn2017model,rakelly2019efficient,jacobson2025hype,10.5555/3546258.3546547}.

\section{Experiments}

\subsection{Setup}

We evaluate whether \zalt can solve novel start--goal pairs in a complex long-horizon environment after seeing only an incomplete set of demonstrations. We ask whether the agent can recombine demonstrated behavioral pieces into new full task solutions without additional online training.

\begin{figure}[t]
    \centering
    \includegraphics[width=1\linewidth]{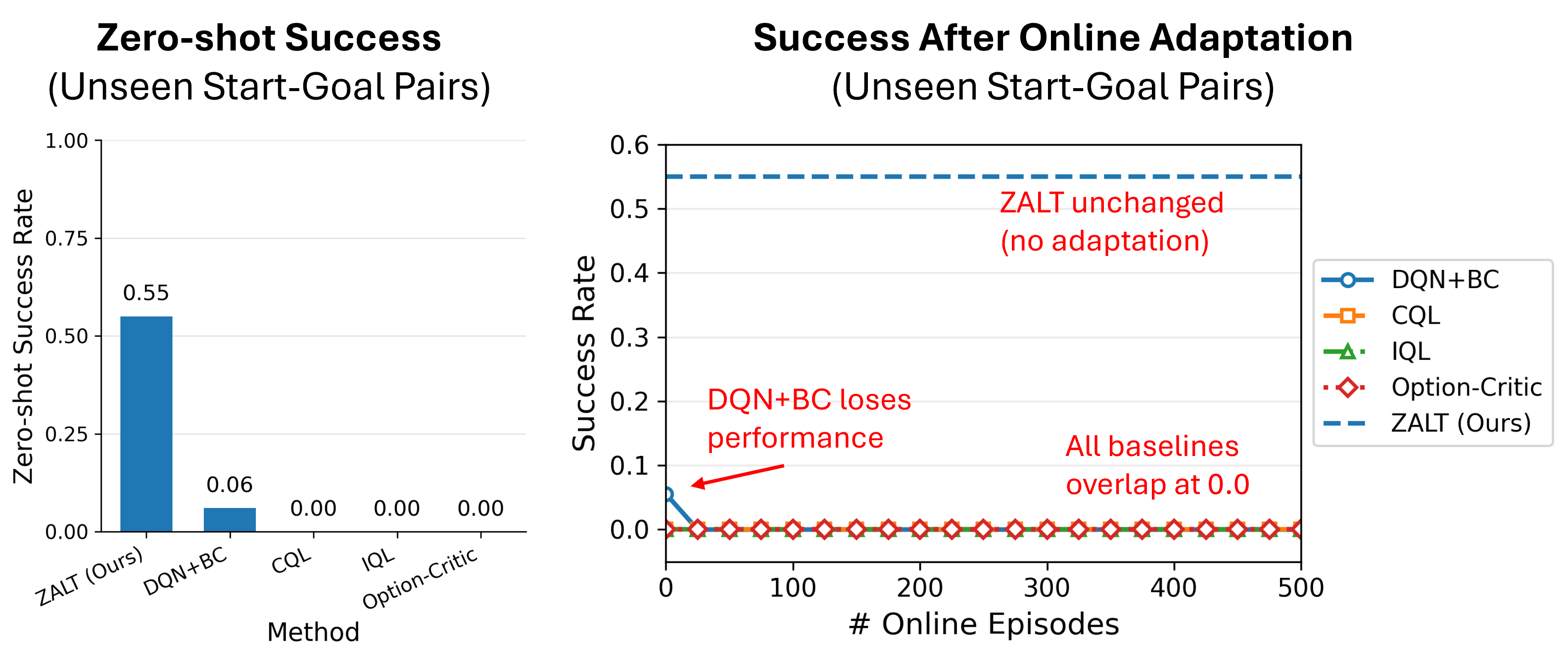}
    \caption{Success rate on unseen start--goal pairs immediately after seeing demonstrations (left), and after a period of online adaptation in the environment (right). \zalt has a zero-shot success rate of 0.55 (immediate success in 55\% of the unseen tasks). The only baseline able to show any zero-shot capability was DQN+BC with a 0.06 success rate. After 500 episodes of adaptation (interacting online with the environment), no baselines gained back performance, likely due to the long horizons.}
    \label{fig:hymap_quant_res}
\end{figure}

\begin{figure}[t]
    \centering
    \includegraphics[width=1\linewidth]{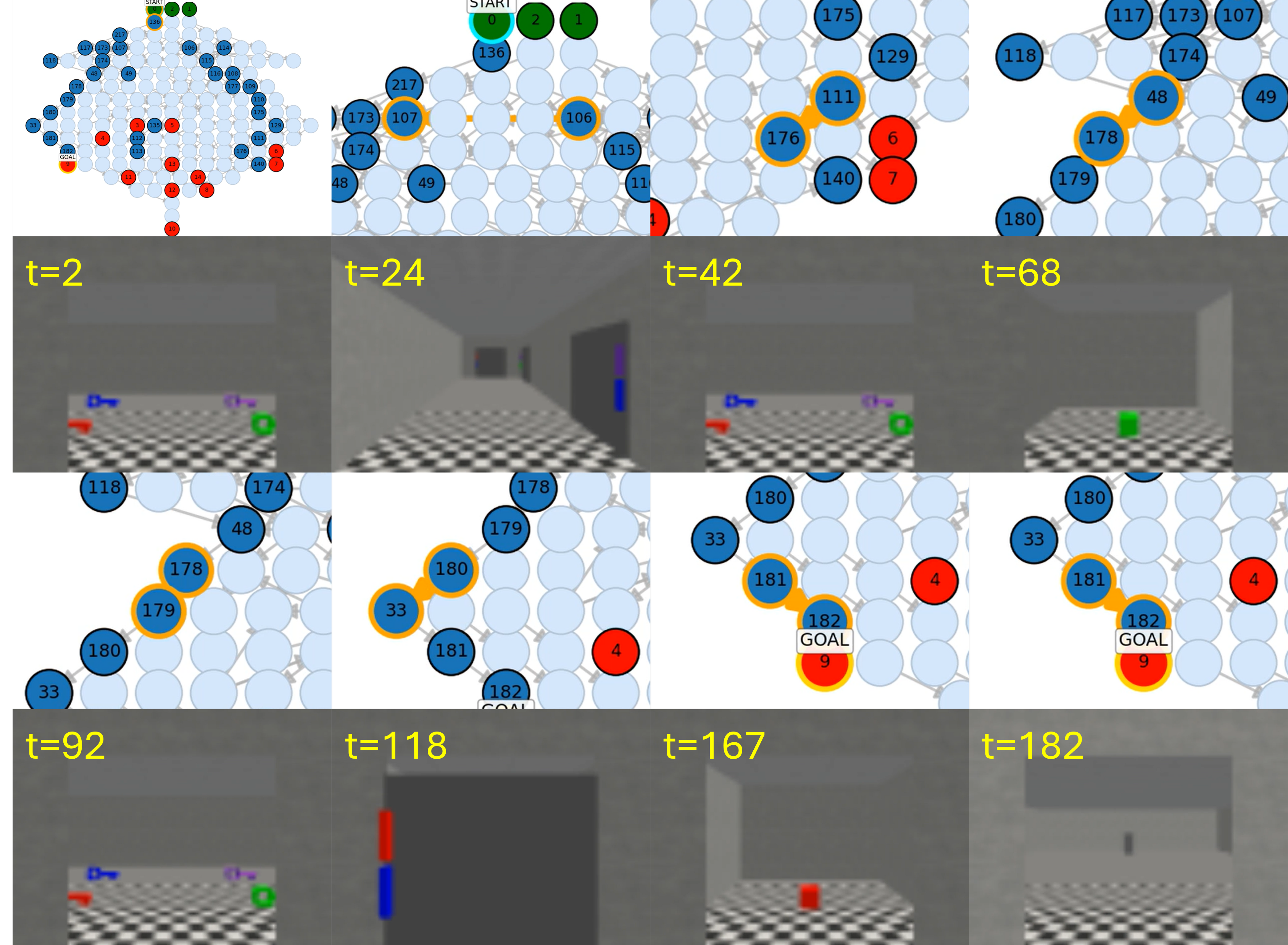}
    \caption{Video frames from a successful \zalt zero-shot inference run (goal: green gem, then red gem). For each step, the top is a visualization of the behavior topology with hubs that are a part of the plan highlighted. Below are the actual images received by the agent. Note that the same visual state (t=2, 42, 92) can have different hubs depending on what has already been done in the environment. In this case, a complex $\sim$200 step traversal was encoded in only 31 hubs. This horizon compression and selection of states to be hubs allowed this task to be completed, despite never appearing in training data (zero-shot). The full video can be viewed at:  \url{https://arcosin.github.io/Zalt/}.}
    \label{fig:hymap_live}
\end{figure}

\noindent\textbf{Environment:} We evaluate in a custom MiniWorld/CARL-style maze environment \cite{chevalierboisvert2023minigrid,benjamins2021carl}. The maze contains one barrel, four colored keys, four locked doors, and four colored diamonds. Each diamond is located behind a locked door, and each door requires a specific pair of keys to open. A goal is an ordered pair of distinct diamond colors: the agent must navigate to the needed keys, unlock the doors for the requested diamonds, collect each diamond, return to the barrel, and deposit the diamonds in the correct order. This makes the task naturally long-horizon and compositional, since many tasks share local behaviors but require different ordered combinations. The goal space contains all $4 \times 3 = 12$ ordered two-diamond goals, and evaluation uses three fixed start configurations, giving 36 total start--goal pairs. Observations include a first-person RGB image and a compact barrel-state vector recording which diamonds have already been deposited. The discrete actions are move forward, move backward, turn left, turn right, pick up, and toggle/interact. Full environment details are provided in Appendix~\ref{sec:appendix_maze}.

Some baselines used require a reward function. For this environment, the reward function is sparse and conditional on the goal: $r(s,a,g) = 1$ when the goal is achieved and a constant small negative reward (e.g. $-0.001$) otherwise.

\noindent\textbf{Demonstrations:} The demonstration dataset contains 18 successful demonstrations and 120 failed demonstrations. Successful demonstrations are intentionally incomplete by start: each of the three start configurations receives demonstrations for only six of the 12 possible ordered goals, giving 18 seen start--goal pairs. The remaining six goals for each start are held out, giving 18 unseen start--goal pairs for zero-shot evaluation. Failed demonstrations are sampled only from canonical mistakes on the seen start--goal subset, such as wrong-key door attempts or incorrect diamond deposits. They do not contain successful trajectories or successful task prefixes for held-out start--goal pairs. They are included to expose unsupported routes and failure endpoints. Full dataset construction details are provided in Appendix~\ref{sec:appendix_demos}.

\noindent\textbf{\zalt Setup:} \zalt uses a 64-dimensional latent state, tolerance $\epsilon=0.001$ for hub matching, and history length 75 for the memory-augmented encoder and hub policies. The low-level latent dynamics model is trained with learning rate $10^{-4}$, the high-level hub dynamics model with learning rate $2 \times 10^{-4}$, and the hub-to-hub behavior-cloning policies with learning rate $10^{-3}$. The high-level model is trained on hub sequences extracted from demonstrations, with topology masking over valid outgoing hub edges. We selected hyperparameters by one-dimensional sweeps over learning rate, hub tolerance, and history length while holding other parameters fixed; full architecture and tuning details are provided in Appendix~\ref{sec:appendix_zalt_setup}.

\noindent\textbf{Baselines:}
We compare against four recurrent, goal-conditioned baselines. All baselines use GRU memory so that hidden carried objects and recent interaction history are not treated as one-frame Markov information. DQN+BC combines a DQN-style temporal-difference objective with a behavior-cloning loss on demonstrated actions \cite{mnih2015human}. CQL is a discrete-action conservative offline RL baseline that penalizes unsupported actions while learning from the fixed demonstration dataset \cite{kumar2020conservative}. IQL is an implicit Q-learning baseline adapted to the same goal-conditioned recurrent setting \cite{kostrikov2022offline}. Option-Critic is a hierarchical baseline with latent options, intra-option policies, and learned termination behavior \cite{bacon2017option,sutton1999between}.  Hyperparameter sweeps and implementation details are provided in Appendix~\ref{sec:appendix_baselines}.

\noindent\textbf{Ablations:}
We evaluate two ablations to test whether \zalt's main components are necessary. First, we replace the high-level hub dynamics model with breadth-first search over the behavior topology. This tests whether explicit hub-level prediction is needed, or whether topology connectivity alone is sufficient. Second, we remove the GRU memory from the low-level encoder--dynamics--decoder model. In this ablation, hub discovery is based only on the current observation, so hubs cannot encode recent behavioral context such as previously collected objects or completed interactions.

\noindent\textbf{Metrics:}
We report task success rate, defined as the fraction of start--goal pairs for which the agent deposits the two requested diamonds in the correct order within the episode horizon. We primarily measure zero-shot success on the 18 held-out start--goal pairs, while also tracking seen-pair success as a sanity check that methods can reproduce demonstrated tasks. Each baseline is first evaluated zero-shot after offline training on the demonstration dataset, and then evaluated again after 500 episodes of online adaptation in the environment. During adaptation, we measure held-out success periodically to test whether additional interaction can recover performance. \zalt is not adapted online in this experiment (it does not define an online adaptation mechanism). Its zero-shot success rate is shown as a fixed reference during the adaptation comparison.

\subsection{Results \& Discussion}

Figure~\ref{fig:hymap_quant_res} (left) shows that \zalt substantially outperforms all baselines on held-out start--goal pairs. \zalt solves 55\% of unseen tasks zero-shot, while DQN+BC solves 6\% and CQL, IQL, and Option-Critic solve 0\%. Since DQN+BC solves only one of the 18 unseen tasks, the baseline results indicate that ordinary goal-conditioned policies do not reliably recombine the demonstrated pieces into new long-horizon routes. This supports the core claim that explicit behavior topology is useful when the full start--goal route is out of distribution but its local behavioral pieces are present in the dataset.

The seen-task results show that the environment is difficult even before zero-shot generalization. \zalt solves 13 of 18 seen tasks, or 72.2\%, while the strongest baseline, DQN+BC, solves 2 of 18, or 11\%. The BC component likely explains why DQN+BC is the strongest baseline: it gives the method direct pressure to imitate demonstrated actions, while CQL and IQL must recover useful long-horizon behavior through value learning from sparse rewards. Option-Critic has a natural connection to temporal abstraction, but its options are latent and not tied to explicit demonstrated junctions or hub-to-hub connectivity, so it does not discover the needed compositional structure.

Diving deeper into \zalt's scores, we found that the average number of edges crossed for seen and unseen tasks was 33.61 and 33.39 respectively. When translated to low-level control, these tasks averaged 243.78 and 281.39 timesteps respectively. On average, an edge encoded for 7.9 primitive actions.  This shows that the behavior topology provides substantial horizon compression: planning occurs over tens of hub transitions rather than hundreds of primitive decisions. This compression is important because it reduces the number of high-level decisions over which errors can accumulate. \zalts successful runs can also be seen visually in Figure~\ref{fig:hymap_live} and its accompanying video. Here, it completes a complex $\sim$200 step traversal by moving over 31 hubs in the behavior topology. 

The online adaptation results in Figure~\ref{fig:hymap_quant_res} (right) show that 500 additional environment episodes are not enough for the baselines to recover held-out performance. DQN+BC loses its small initial success, and CQL, IQL, and Option-Critic remain at 0\%. \zalt is unchanged during this adaptation comparison, so its 55\% line represents the zero-shot performance of the learned behavior topology rather than improvement from online interaction. This suggests that the challenge is not that the baselines need a small amount of extra practice. The task has sparse rewards, long horizons, local reward maxima, and many ways to fail before reaching the goal.

The ablations show that both components are necessary for \zalt to work in this environment. Replacing the high-level hub dynamics model with breadth-first search over the topology gives 0\% success on both seen and unseen tasks. This indicates that raw topology connectivity is not enough: the planner must account for which hub transitions are plausible given the previous hub history. In this ablation, the pure topology attempted routes that were impossible, such as opening the blue lock with the red key. Removing the GRU memory from the low-level encoder--dynamics--decoder model also gives 0\% success on both seen and unseen tasks. This indicates that hub discovery cannot rely only on the current observation. Recent behavioral context is needed to distinguish states that look similar but differ in task-relevant history. A wider discussion of \zalts assumptions, limitations, and possible extensions is provided in Appendix~\ref{sec:appendix_limitations}.

\section{Conclusion}

We introduced \zalt, an imitation-learning method for solving unseen long-horizon start--goal tasks by recombining behavior from an incomplete demonstration dataset. \zalt converts demonstrations into a latent behavior topology, identifies hub states where trajectories converge or diverge, learns a high-level dynamics model over hub sequences, and executes planned hub routes with demonstration-trained hub-to-hub policies. In a complex 3D maze task, this explicit topology enables 55\% zero-shot success on held-out start--goal pairs, substantially outperforming baselines. Future work will study the underlying forces that cause useful hubs to form in long-horizon environments and adapt \zalt to online or hybrid online-demonstration learning settings.

\FloatBarrier
\bibliographystyle{plain}
\bibliography{main}

\FloatBarrier
\appendix
\newpage

\section{Method Additional Material}
\label{sec:appendix_more_maze}

\subsection{Networks Summary}

\zalt learns the following parameterized neural networks:

\begin{center}
\small
\begin{tabular}{lll}
$Enc_{\theta_e}(s_t,\text{history}_t) \rightarrow z_t$
&
$M^\text{low}_{\theta_m}(z_t,a_t) \rightarrow \hat{z}_{t+1}$
&
$Dec_{\theta_d}(\hat{z}_{t+1}) \rightarrow \hat{s}_{t+1}$
\\[1.2em]
$M^\text{high}_{\theta_h}(h_{0:m}) \rightarrow P(h_{m+1}\mid h_{0:m})$
&
$\pi_{\theta_\pi,h_i}(s_t,h_j,\text{history}_t) \rightarrow P(a_t \mid s_t,h_j)$
&
\end{tabular}
\end{center}

These are trained using the following losses:

\begin{center}
\small
\begin{tabular}{ll}
\textbf{Low-level latent dynamics} &
$\mathcal{L}_{low}
= w_z \mathcal{L}_{z}
+ w_{vis}\mathcal{L}_{vis}$
\\[0.8em]

\textbf{Latent prediction} &
$\mathcal{L}_{z}
=
\|M^\text{low}_{\theta_m}(z_t,a_t)
-
Enc_{\theta_e}(s_{t+1},\text{history}_{t+1})\|^2$
\\[0.8em]

\textbf{Visual reconstruction} &
$\mathcal{L}_{vis}
=
\|Dec_{\theta_d}(M^\text{low}_{\theta_m}(z_t,a_t))^{vis}
-
s^{vis}_{t+1}\|^2_{\mathrm{weighted}}$
\\[0.8em]

\textbf{High-level hub dynamics} &
$\mathcal{L}_{high}
=
-\log M^\text{high}_{\theta_h}(h_{m+1}\mid h_{0:m})$
\\[0.8em]

\textbf{Hub policies} &
$\mathcal{L}_{BC}
=
-\log \pi_{\theta_\pi,h_i}(a_t\mid s_t,h_j,\text{history}_t)$
\end{tabular}
\end{center}

\subsection{Low-Level Latent Dynamics Model for Hub Forming}
\label{sec:appendix_low_level_dynamics}
\FloatBarrier

\zalt learns hubs in a latent state space produced by a low-level encoder--dynamics--decoder model over primitive actions. The encoder maps the current observation and recent history into a compact latent state $z_t$. The low-level dynamics model predicts the next latent state after a primitive action, $M^\text{low}_{\theta_m}(z_t,a_t) \rightarrow \hat{z}_{t+1}$. The decoder reconstructs next-state information from the predicted latent, $Dec_{\theta_d}(\hat{z}_{t+1}) \rightarrow \hat{s}_{t+1}$. This follows the general latent-dynamics and world-model approach used in model-based reinforcement learning \cite{ha2018world,hafner2019planet,hafner2020dreamer}.

In our implementation, the encoder is memory-augmented. A feedforward encoder compresses the current high-dimensional observation into an immediate latent summary, while a GRU summarizes recent latent states and adds a history-based correction. These components produce the final latent state, written as $Enc_{\theta_e}(s_t,\text{history}_t) \rightarrow z_t$. During training, history is handled with clipped backpropagation through time. This improves stability and limits the ability of the latent state to memorize full trajectory identity, encouraging behaviorally similar visits to the same junction to merge in latent space.

The model is trained with a weighted combination of latent prediction and reconstruction losses. The latent prediction loss is $\mathcal{L}_{dyn} = \| M^\text{low}_{\theta_m}(z_t,a_t) - Enc_{\theta_e}(s_{t+1},\text{history}_{t+1}) \|^2$. The reconstruction loss is $\mathcal{L}_{rec} = \| Dec_{\theta_d}(M^\text{low}_{\theta_m}(z_t,a_t)) - s_{t+1} \|^2$. The reconstruction term includes the next observation information needed for the task, including the first-person visual state and task-progress state. The total training objective is a weighted sum of these terms, with the model used to produce a dynamics-aware representation for hub discovery rather than for long-horizon rollout planning.

\subsection{\zalt Inference Details}
\label{sec:appendix_inference}
\FloatBarrier

At inference time, \zalt searches over hub histories rather than single hub states, because the plausibility of a transition can depend on the route used to reach the current hub. Each partial route $H=(h_0,\ldots,h_m)$ is expanded using the high-level model $M^\text{high}_{\theta_h}$, with invalid topology edges masked out and very low-probability transitions rejected. Routes are scored with the bottleneck cost $C(H')=\max(C(H),-\log P(h'\mid H))+\eta$. Intuitively, a plan is only as reliable as its weakest transition: one highly implausible hub jump can make the whole route fail, even if the other transitions are likely. The small hop penalty $\eta$ breaks ties in favor of shorter routes when bottleneck quality is similar.

Algorithm~\ref{alg:hub_history_search} gives the resulting search procedure. The queue stores complete hub histories and repeatedly expands the lowest-cost history. If the current hub is a goal hub, the history is returned as the plan. Otherwise, the model predicts the next-hub distribution conditioned on the full history, unsupported outgoing edges are masked, and each remaining neighbor above the probability floor is inserted back into the queue with its updated bottleneck cost. If no goal-reaching history is found within the depth limit, inference returns failure.

\begin{algorithm}[tbh]
\caption{Hub-History Search with $M^\text{high}$}
\label{alg:hub_history_search}
\begin{algorithmic}[1]
\Require Topology $G=(\mathcal{H},\mathcal{E})$, high-level model $M^\text{high}_{\theta_h}$, start hub $h_s$, goal hubs $\mathcal{H}_g$, hub-depth limit $D$
\State Initialize queue with history $H=(h_s)$ and cost $C(H)=0$
\While{queue is not empty}
    \State Remove the lowest-cost history $H=(h_0,\ldots,h_m)$
    \If{$h_m \in \mathcal{H}_g$}
        \State \Return hub plan induced by $H$
    \EndIf
    \If{$|H|-1 = D$}
        \State continue
    \EndIf
    \State Use $M^\text{high}_{\theta_h}(H)$ to compute $P(h' \mid H)$ over next hubs
    \State Mask all hubs $h'$ where $(h_m,h') \notin \mathcal{E}$
    \For{each remaining outgoing hub $h'$}
        \If{$P(h' \mid H) < p_{\min}$}
            \State continue
        \EndIf
        \State Extend the history: $H'=(h_0,\ldots,h_m,h')$
        \State Score it by bottleneck cost: $C(H')=\max(C(H),-\log P(h'\mid H))+\eta$
        \State Insert or update $H'$ in the queue with priority $C(H')$
    \EndFor
\EndWhile
\State \Return failure
\end{algorithmic}
\end{algorithm}

\FloatBarrier

\section{Maze Game Specification}
\label{sec:appendix_maze}
\FloatBarrier

We evaluate \zalt in a custom MiniWorld-based 3D maze environment designed to require long-horizon, ordered object manipulation. The environment contains one barrel, four colored keys, four locked doors, and four colored diamonds. Each diamond is located behind a locked door, and each door requires a fixed pair of key colors to open. The agent must collect keys (one held at a time), unlock the relevant doors, pick up diamonds, return to the barrel, and deposit two diamonds in the requested order.

The maze is grid-snapped for controlled long-horizon behavior. The agent moves in cell-sized steps and rotates in $90^\circ$ increments. The discrete action space has six actions: move forward, move backward, turn left, turn right, pick up, and toggle/interact. Toggle is used both for applying keys to doors and depositing diamonds into the barrel. The main evaluation horizon is capped at 400 primitive steps, which is almost double the number of steps needed for the longest tasks.

The object layout is deterministic. The barrel is located in the main room. The four keys are placed in a key room, with one red, blue, green, and purple key. The four diamonds are placed behind four different locked doors. The red diamond door requires red and blue keys, the blue diamond door requires red and green keys, the green diamond door requires blue and purple keys, and the purple diamond door requires green and purple keys. Key order is not important for the environment: a door opens after both required keys have been used. After the first correct key is used, the door is partially opened; after the second correct key is used, the door is removed and the passage becomes traversable. Keys are consumed when used, and picked-up keys respawn at their original key-room locations so the same key color can be reused later in the task. Wrong-diamond deposits and wrong-key door attempts are terminal in the environment without success.

A task goal is an ordered pair of distinct diamond colors. With four diamond colors, this gives $4 \times 3 = 12$ possible ordered goals. A task succeeds only when the deposited barrel sequence exactly matches the requested two-diamond order. Depositing the wrong diamond terminates the episode as a failure. The environment reward uses a small step cost of $-0.1$, a success reward of $100$, and a failure penalty for wrong deposits.

Observations contain a first-person RGB image of shape $60 \times 80 \times 3$, and a length-two barrel-state vector. The barrel-state vector records the diamonds already deposited using $0$ for empty and integer color IDs for red, blue, green, and purple. Held keys and held diamonds are hidden from the rendered first-person image, so the agent must rely on its learned state and memory rather than directly seeing the carried object in front of the camera.

Evaluation uses three fixed start configurations: $\sigma_A=((3,3),0)$, $\sigma_B=((3,6),0)$, and $\sigma_C=((6,3),0)$, where the first pair gives grid position and the second value gives initial orientation. Since there are 12 ordered two-diamond goals and three fixed starts, the full evaluation space contains $3 \times 12 = 36$ start--goal pairs. 


\section{Demonstration Dataset Specification}
\label{sec:appendix_demos}
\FloatBarrier

The demonstration dataset uses the three fixed start configurations and the 12 ordered two-diamond goals described in Appendix~\ref{sec:appendix_maze}. Successful demonstrations are intentionally incomplete by start: each start is paired with six of the 12 goals, giving 18 successful training start--goal pairs. The remaining six goals for each start are held out, giving 18 unseen start--goal pairs for zero-shot evaluation. The successful-goal assignment is deterministic and overlapping: if the ordered goal list is $(g_0,\ldots,g_{11})$, then start 0 receives $g_0,\ldots,g_5$, start 1 receives $g_4,\ldots,g_9$, and start 2 receives $g_8,\ldots,g_{11},g_0,g_1$. Thus, each start is missing half of the goals, while every goal appears in at least one successful demonstration from some start.

Each successful demonstration completes the requested ordered goal exactly. For a goal $(g_0,g_1)$, the expert collects the two required keys for $g_0$, unlocks the corresponding door, collects diamond $g_0$, deposits it in the barrel, and then repeats the same process for $g_1$. In the main dataset, we collect one successful demonstration for each seen start--goal pair, for 18 successful demonstrations total. 

The dataset also includes failed demonstrations to expose incorrect routes and unsupported behavior. Failure specifications are generated only for the seen goal subset of each start, so they do not contain successful held-out start--goal behavior. Failures include using an incorrect key on a door, depositing the second requested diamond first, depositing an unrelated diamond first, and depositing an unrelated diamond after a correct first deposit. There are 13 canonical failure specifications per seen ordered goal, giving $18 \times 13 = 234$ possible start/failure candidates. We sample 120 failed demonstrations uniformly without replacement.

Demonstrations are generated by an expert planner over a discrete grid abstraction of the environment and then executed in the original 3D environment to record transition tuples. 


\section{\zalt Implementation Details}
\label{sec:appendix_zalt_setup}
\FloatBarrier

\zalt uses a 64-dimensional latent state, hub tolerance $\epsilon=0.001$, and history length 75. The low-level model has an image/vector encoder, a GRU memory module, a latent dynamics predictor, and a decoder for next-observation information. It is trained for 300 epochs with learning rate $10^{-4}$ using a weighted combination of latent prediction, visual reconstruction, barrel-vector reconstruction, and terminal-status losses.

Hub extraction is performed by tolerance-bucketing latent states and identifying starts, terminal states, convergences, and divergences as described in the main text. We selected $\epsilon=0.001$ by sweeping 17 logarithmically spaced tolerances from $10^{-4}$ to $1.0$ and inspecting the resulting hub counts and nearest-neighbor distances. History length was selected from $\{0,2,10,25,50,75,100,150,200\}$. Learning rates were selected from sweeps over $10^{-4}$, $3 \times 10^{-4}$, and $10^{-3}$ while holding other parameters fixed.

The high-level hub dynamics model is a GRU over hub embeddings followed by a next-hub classifier with topology masking. It is pretrained on random valid hub-graph traversals, then trained on demonstrated hub sequences using learning rate $2 \times 10^{-4}$. Hub policies use one target-conditioned policy per source hub and are trained by behavior cloning with learning rate $10^{-3}$. Boundary perturbation uses canonical segments with probability 0.8, truncated segments with probability 0.1, and preroll segments with probability 0.1; the maximum perturbation length is 3 steps. We also use observation noise of 0.01 and action-label smoothing of 0.05.

\section{Baseline Implementation Details}
\label{sec:appendix_baselines}
\FloatBarrier

All baselines are goal-conditioned recurrent agents trained from the same demonstration dataset as \zalt. Each uses a visual encoder followed by a GRU memory module and is trained on contiguous trajectory chunks with truncated backpropagation through time. This gives the baselines access to recent history, which is important because held keys and diamonds are not directly visible in the rendered observation.

We evaluate four baselines. DQN+BC uses a recurrent Q-network trained with a weighted sum of a DQN temporal-difference loss and a behavior-cloning loss. CQL uses recurrent goal-conditioned Q-functions with the standard conservative penalty over discrete actions. IQL uses recurrent goal-conditioned Q, value, and policy heads with expectile value learning and advantage-weighted policy learning. Option-Critic uses a recurrent shared encoder with latent options, intra-option action policies, and learned option termination.

For hyperparameter selection, we used local grid sweeps that vary learning rate and the main method-specific parameter while holding the rest of the configuration fixed. DQN+BC sweeps learning rate and behavior-cloning weight. CQL sweeps learning rate and conservative penalty weight. IQL sweeps learning rate, expectile, and policy advantage temperature. Option-Critic sweeps learning rate, number of options, and termination weight. The shared sequence length is 75 and the shared GRU hidden size is 64 in the sweep configuration.

After offline training, each baseline is evaluated on the seen and held-out start--goal cases. We also evaluate whether online interaction can recover performance after the zero-shot setting. For this adaptation experiment, each baseline starts from its offline checkpoint and receives 500 live environment episodes, with 25 gradient updates per episode from recurrent replay. Evaluation is performed periodically during adaptation and finally after the 500-episode adaptation budget.

\section{Limitations}
\label{sec:appendix_limitations}

\zalt assumes that demonstrations contain reusable behavioral connectivity. This is a natural assumption for many long-horizon imitation-learning settings, since expert demonstrations often share rooms, objects, tools, subgoals, or navigation corridors even when the complete start--goal task differs. When this connectivity exists, \zalt is designed to expose it explicitly through hubs and hub-to-hub policies. If a required segment is absent, the topology correctly has no supported edge for that behavior, avoiding unsupported primitive-action extrapolation.

A second limitation is that hub quality depends on the learned latent representation. If the latent space groups states by superficial visual similarity rather than behavioral similarity, the resulting hubs may be less useful. \zalt mitigates this by learning the representation with a low-level dynamics objective rather than pure reconstruction alone, encouraging latent states to reflect what actions can do next. The GRU-based encoder further helps separate visually similar states with different histories, such as states where the agent appears near the same door but has collected different keys. Nonetheless, poor hub quality can lead to poor learning of hub-to-hub policies or difficult planning.

Inference also requires search over hub histories, so the number of possible histories can grow with topology size and branching factor. In practice, \zalt mitigates this through topology masking, probability-floor pruning, a hub-depth limit, and a bottleneck transition cost that discourages implausible routes. This makes search focus on supported, high-probability hub transitions rather than arbitrary paths through the graph. Moreover, the purpose of the topology is to replace hundreds of primitive decisions with a much smaller number of abstract hub transitions, so the added high-level search cost is traded for a large reduction in primitive-horizon complexity.

Finally, \zalt currently builds a fixed topology from a fixed demonstration dataset. This is appropriate for the zero-shot setting studied here, where no additional interaction or online optimization is allowed. In future online or hybrid settings, the same framework could naturally be extended by adding newly observed hubs, updating edge policies, or refining the high-level dynamics model as more experience becomes available.

\section{Broader Impacts}
\label{sec:appendix_broader_impacts}

This work studies zero-shot imitation learning in a simulated 3D maze environment. There is no direct path from the presented system to negative real-world applications. Indirectly, improved imitation learning and long-horizon behavior composition could contribute to more capable autonomous agents and robotics systems, with potential positive uses in safer task learning, reduced data collection burden, and improved generalization from demonstrations. The same general capabilities could also be misapplied if transferred into real-world autonomous systems without appropriate task constraints, monitoring, or safety validation. In this paper, the method is evaluated only in a controlled simulated environment and does not provide a deployed real-world control system.

\section{Computational \& Software Resources}
\label{sec:appendix_resources}
All experiments were run using PyTorch with CUDA acceleration on a Windows 11 machine with an Intel64 Family 6 Model 198 CPU and an NVIDIA GeForce RTX 5080 Laptop GPU with 16 GB GPU memory.


\end{document}